\documentclass{article}
\usepackage{spconf,amsmath,graphicx}

\usepackage{url,amsfonts,amssymb,bm}

\usepackage{enumitem}
\usepackage{algorithm}
\usepackage{algorithmicx, algpseudocode }
\usepackage{multirow}
\usepackage{color}
\usepackage{mathtools}
\usepackage{caption}
\usepackage{subcaption}
\usepackage{amsthm}

\usepackage[colorlinks=true, linkcolor=blue, citecolor=blue, urlcolor=blue]{hyperref}
\hypersetup{pdfcreator  = {LaTeX with hyperref package},
               pdfproducer = {dvips + ps2pdf} }

\usepackage{tikz}
\usetikzlibrary{positioning}

\theoremstyle{definition}

\DeclareMathOperator*{\diag}{diag}

\DeclareMathOperator*{\unique}{unique}

\newcommand{\bdiag}{{\hbox{Bdiag}}}
\newcommand{\floor}{{\hbox{floor}}}

\usepackage{bm}

\long\def\comment#1{}

\newfont{\bbb}{msbm10 scaled 700}

\newfont{\bb}{msbm10 scaled 1100}

\title{Region adaptive graph fourier transform for 3d  point clouds }
\name{Eduardo Pavez$^\star$, Benjamin Girault$^\star$, Antonio Ortega$^\star$, Philip A. Chou$^\dagger$ \thanks{ Author email: pavezcar@usc.edu.  This work was funded in part by NSF under grant CCF-1410009, and by a Google Faculty Research Award.}}
\address{$^\star$University of Southern California, Los Angeles, California, USA \\
$^\dagger$Google, Inc., Mountain View, California, USA}
\begin{document}
\ninept
\maketitle
\begin{abstract}
We introduce the Region Adaptive Graph Fourier Transform (RA-GFT) for compression of 3D point cloud attributes.
The RA-GFT is a multiresolution   transform, formed by combining   spatially localized block transforms.  We assume  the points are organized by a family of nested partitions represented by a rooted tree.   At each resolution level,   attributes are processed in clusters using  block transforms. Each block transform produces a single approximation (DC) coefficient, and various detail (AC) coefficients. The DC coefficients are promoted up the tree to the next (lower resolution) level, where the process can be repeated until reaching the root. Since  clusters may have a different numbers of points, each block transform must incorporate the relative importance of each coefficient. For this, we introduce the $\mathbf{Q}$-normalized graph Laplacian, and propose using its eigenvectors as the block transform. The RA-GFT achieves better complexity-performance trade-offs than previous approaches. In particular, it  outperforms the  Region Adaptive Haar Transform (RAHT) by up to 2.5 dB, with a small   complexity overhead. 
\end{abstract}
\section{Introduction}
\label{sec:intro}
Driven by applications in virtual and augmented reality,  remote sensing, and autonomous vehicles, 
it is now possible to capture, in real time and at low cost, time varying 3D scenes, public spaces with moving objects, and people. The preferred representation for such data are 3D point clouds, which consist of 1) a list of \textit{3D point coordinates}, and 2) \textit{attributes} associated with those coordinates, such as color. 
In many applications, large point clouds need to be compressed for storage and transmission, leading to the recent development of a  standard by the moving pictures expert group (MPEG)  \cite{schwarz2018emerging}. 

We present a new transform for attribute compression, which often takes up more than half of the overall bit budget for typical point clouds. 
Motivated by transforms used in image, video and point cloud compression \cite{vetterli_wavelet_1995,ahmed1974discrete,queiroz2016compression}, we construct our transform with the goal of achieving three fundamental properties:   1) orthogonality, 2) a constant basis function, and 3)   low complexity. Orthogonality ensures that errors in the transform and point cloud attribute domains are equal. 
A constant basis function guarantees that an attribute with the same value at all points  (the smoothest signal) has the most compact representation ($1$-sparse signal) in the transform domain. Finally,  for the transform to scale to point clouds with a large number of points, $N$, we require a complexity of $\mathcal{O}(N\log N)$, instead of a naive implementation that uses matrix vector products, which would require $\mathcal{O}(N^2)$ complexity, if the transform is available explicitly, or $\mathcal{O}(N^3)$, if it has to be obtained via eigendecomposition.  
%

\begin{figure}[ht]
    \centering
    \begin{tikzpicture}[scale=0.8]
        \node[blue!40!white,draw,circle,inner sep=1pt,fill,text=black] (v1) at (0, 0) {$\mathbf{v}_1$};
        \node[blue!40!white,draw,circle,inner sep=1pt,fill,text=black] (v2) at (1, 0) {$\mathbf{v}_2$};
        \node[blue!40!white,draw,circle,inner sep=1pt,fill,text=black] (v3) at (2, 0) {$\mathbf{v}_3$};
        \node[blue!40!white,draw,circle,inner sep=1pt,fill,text=black] (v4) at (3, 0) {$\mathbf{v}_4$};
        \node[blue!40!white,draw,circle,inner sep=1pt,fill,text=black] (v5) at (4, 0) {$\mathbf{v}_5$};
        \node[blue!40!white,draw,circle,inner sep=1pt,fill,text=black] (v6) at (5, 0) {$\mathbf{v}_6$};
        \node[blue!40!white,draw,circle,inner sep=1pt,fill,text=black] (v7) at (6, 0) {$\mathbf{v}_7$};
        \node[blue!40!white,draw,circle,inner sep=1pt,fill,text=black] (v8) at (7, 0) {$\mathbf{v}_8$};
        \node[blue!40!white,draw,circle,inner sep=1pt,fill,text=black] (v9) at (8, 0) {$\mathbf{v}_9$};
        
        \node[anchor=west] (l3) at (8.5,0) {\footnotesize $\ell=L=2$};
        
        \node[green!50!white,draw,circle,inner sep=1pt,fill,text=black] (V1_1) at (1,1) {$\mathbf{v}_1^1$};
        \draw (v1) -- (V1_1);
        \draw (v2) -- (V1_1);
        \draw (v3) -- (V1_1);
        
        \node[green!50!white,draw,circle,inner sep=1pt,fill,text=black] (V2_1) at (4.5,1) {$\mathbf{v}_2^1$};
        \draw (v4) -- (V2_1);
        \draw (v5) -- (V2_1);
        \draw (v6) -- (V2_1);
        \draw (v7) -- (V2_1);
        
        \node[green!50!white,draw,circle,inner sep=1pt,fill,text=black] (V3_1) at (7.5,1) {$\mathbf{v}_3^1$};
        \draw (v8) -- (V3_1);
        \draw (v9) -- (V3_1);
        
        \node[anchor=west] (l3) at (8.5,1) {\footnotesize $\ell=1$};
        
        \node[orange!80!white,draw,circle,inner sep=1pt,fill,text=black] (V1_0) at (4,2) {$\mathbf{v}_1^0$};
        \draw (V1_1) -- (V1_0);
        \draw (V2_1) -- (V1_0);
        \draw (V3_1) -- (V1_0);
        
        \node[anchor=west] (l3) at (8.5,2) {\footnotesize $\ell=0$};
    \end{tikzpicture}
    \caption{Nested partition   represented by a  hierarchical tree.
    Each level of the tree partitions the nodes in the next level.  Thus at level $\ell=1$, $\mathbf{v}_1^1$, $\mathbf{v}_2^1$, and $\mathbf{v}_3^1$ respectively represent the sets $\mathcal{V}_1^1=\{\mathbf{v}_1,\mathbf{v}_2,\mathbf{v}_3\}$, $\mathcal{V}_2^1=\{\mathbf{v}_4,\mathbf{v}_5,\mathbf{v}_6,\mathbf{v}_7\}$, and $\mathcal{V}_3^1=\{\mathbf{v}_8,\mathbf{v}_9\}$, while at level $\ell=0$, $\mathbf{v}_1^0$ represents the set $\mathcal{V}_1^0=\{\mathbf{v}_1^1,\mathbf{v}_2^1,\mathbf{v}_3^1\}$.
    } 
    \label{fig:tree_cluster}
\end{figure}
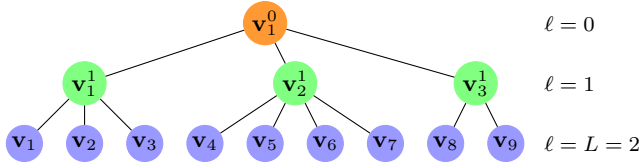

We propose the \emph{Region Adaptive Graph Fourier Transform (RA-GFT)}, a multiresolution transform formed by  combining spatially localized block transforms, where each block corresponds to a cube in 3D space.  The points 
are organized as a set of nested  partitions represented by a tree. Leaf nodes represent points in the original point cloud, while each internal node represents all points within the corresponding subtree  (see Fig.~\ref{fig:tree_cluster}). 
Resolution levels are determined  by the levels of the tree, with higher resolution corresponding to the deepest level (representing single points), and the coarsest resolution corresponding to the root (representing all points). At each resolution level, attributes belonging to the same group  (points that have the same parent in the tree)  are passed through an orthogonal transform for decorrelation. Each block transform produces a single approximation (DC) coefficient, and several detail (AC) coefficients. The DC coefficients are promoted to a lower resolution level where the same process can be repeated until reaching the root\footnote{This process can also stop before reaching the root, so that multiple subtrees, each with its own DC coefficient, are stored. This may be more efficient for large point clouds where there is limited correlation across subtrees.  }.  
%
%

Since each internal node in the tree represents a set, possibly containing a different number of points, the  block transforms   should incorporate the relative importance of the nodes, based on their respective number of descendants. To address this issue, we propose a new graph Fourier transform (GFT) given by the eigenvectors of the $\mathbf{Q}$-normalized graph Laplacian $\mathbf{L_Q} =\mathbf{Q}^{-1/2}\mathbf{L}\mathbf{Q}^{-1/2} $, where $\mathbf{Q}$ is a diagonal matrix with diagonal entries containing the number of descendants of each node, and $\mathbf{L}$ is the combinatorial graph Laplacian. In contrast to the normalized or combinatorial Laplacian matrices \cite{shuman_emerging_2013}, our new variation operator encodes  the local geometry (distances between points) in $\mathbf{L}$ as well as the relative importance of a given set of points.  The proposed transform is closely related to the Irregularity Aware Graph Fourier Transform (IAGFT) \cite{girault2018irregularity}. Our code is available at: \url{https://github.com/STAC-USC/RA-GFT}.

Multiresolution decompositions for point cloud coding have been proposed  based on  graph filter bank theory \cite{anis2016compression,thanou2016graph, chou2019volumetric}. However, they often lack orthogonality,  and the  multiresolution representations are built through complex graph partitioning and reduction algorithms, which make them impractical for large point clouds. 
A Haar-like basis was proposed in \cite{gavish2010multiscale} for any data that can be represented by a hierarchical tree. This construction is orthogonal and  has a constant basis function. Although the other basis functions are spatially localized, they do not exploit  local geometry information (distances between points). In addition, there is no efficient algorithm for computing transform coefficients, and matrix vector products need to be used. 
Also inspired by the Haar transform, \cite{tremblay2016subgraph} proposed sub-graph based filter banks, where a graph is partitioned into connected sub-graphs. For each sub-graph, a local Laplacian based GFT  is used. 
 Although \cite{tremblay2016subgraph} and the RA-GFT follow a similar strategy based on  nested partitions, the  graph construction,   transforms,  coefficient arrangement, and design goals are different. 

 The RA-GFT can be applied to any type of dataset as long as a nested  partition is available. For 3D point clouds,  a natural choice is the octree decomposition \cite{jackins1980oct,meagher1982geometric}, which   can be used to  implement RA-GFT for point clouds with $\mathcal{O}(N\log N)$ complexity.  This data structure has already been used to design transforms for point cloud attributes \cite{queiroz2016compression,zhang2014point,cohen2016attribute,cohen2016point, pavez2018dynamic}. In the block based graph Fourier transform (block-GFT)  \cite{zhang2014point},  the voxel space is partitioned into small blocks, a graph is constructed for the points in each block, and the corresponding graph Fourier (GFT) transform is used to represent attributes in the block.  
 Another popular approach is RAHT \cite{queiroz2016compression}, which is a multiresolution transform  formed by the composition of $2 \times 2$ orthogonal transforms.  
 The block-GFT can achieve excellent performance if the block size is large enough (more points per block), but this has a significant computational cost, since it requires computing GFTs of graphs with possibly hundreds of points. On the other hand, RAHT has an extremely fast implementation, with a competitive coding performance. 
 Our proposed RA-GFT combines ideas from block-GFT and RAHT: it generalizes the block-GFT approach to multiple levels, while RAHT can be viewed as a special case of RA-GFT that is separable and  uses only $2 \times 2 \times 2$ blocks. 
  We demonstrate through point cloud attribute compression experiments that  when RA-GFT is implemented with small block transforms, it can outperform RAHT by up to 2.5dB, with a small computational overhead. When the RA-GFT is implemented with larger blocks, we outperform the block-GFT with a negligible complexity overhead.
  

The rest of this paper is organized as follows. 
Section \ref{sec:ragft} introduces the RA-GFT for arbitrary datasets, while Section \ref{sec:pointclouds} explains how to implement it for 3D point clouds. Compression experiments and conclusions are presented in Sections \ref{sec_exp} and \ref{sec_conclusion} respectively.
\section{Region Adaptive Graph Fourier Transform}
\label{sec:ragft}
\subsection{Notation and preliminaries}
\label{sec:preli}
We use lowercase normal (e.g. $\alpha, a$), lowercase bold (e.g. $\mathbf{v}$) and uppercase bold (e.g.  $\mathbf{A}$) for scalars, vectors and matrices respectively. Vectors and matrices may also be denoted using their entries as $\mathbf{x}=[x_i]$, or $\mathbf{A}=[a_{ij}]$.
Let  $\mathcal{G}=(\mathcal{V,E}, \mathbf{W})$ denote a weighted undirected graph  with  vertex set $\mathcal{V}$, edge set $\mathcal{E}$ and edge weight matrix $\mathbf{W \geq 0}$. An edge weight is positive, that is  $w_{ij} = w_{ji} >0$ if and only if    $(i,j) \in \mathcal{E}$. The graph has $\vert \mathcal{V}\vert =n$ nodes. 
Let $\mathbf{D}=\diag(d_i)$, with $d_i=\sum_j w_{ij}$, be the $n \times n$  degree matrix and let $\mathbf{L}=\mathbf{D}-\mathbf{W}$ be the $n \times n$  combinatorial graph Laplacian matrix. $\mathbf{L}$ is symmetric positive semidefinite, and has eigendecomposition $\mathbf{L}=\mathbf{\Phi \Lambda \Phi^{\top}} $, where the eigenvalues matrix is $\mathbf{\Lambda}=\diag(\lambda_1, \cdots, \lambda_n)$, and the eigenvalues are $\lambda_1=0 \leq \lambda_2, \cdots, \leq \lambda_n$. For connected graphs, $\lambda_2 >0$. The eigenvector associated to $\lambda_1$ is $(1/\sqrt{n})\mathbf{1}$. 
\subsection{RA-GFT block transform}
The Region-Adaptive Graph Fourier Transform (RA-GFT) is an orthonormal transform formed 
from the composition of smaller  block transforms.  
We start by describing the latter.
Let $\mathcal{G}=(\mathcal{V},\mathcal{E},\mathbf{W},\mathbf{Q})$ denote a graph 
as defined in Section \ref{sec:preli}. In addition, define the $n\times n$ \textit{node weight matrix} $\mathbf{Q}=\diag(q_i)$ ($q_i>0$),  the $\mathbf{Q}$-normalized Laplacian $\mathbf{L}_\mathbf{Q}$,  and its eigendecomposition are given by
\begin{equation}
    \mathbf{L}_\mathbf{Q}
    =\mathbf{Q}^{-1/2}\mathbf{L}\mathbf{Q}^{-1/2}
    =\mathbf{\Phi}\mathbf{\Lambda}\mathbf{\Phi}^\top,
\end{equation}
where $\mathbf{\Phi}$ is the matrix of eigenvectors of $\mathbf{L}_\mathbf{Q}$ and $\mathbf{\Lambda}=\diag(\lambda_i)$ is the matrix of eigenvalues.  Since $\mathbf{L}_\mathbf{Q}$ is symmetric and positive semi-definite, $\lambda_i\geq 0$ and $\mathbf{\Phi}$ is orthonormal.  Moreover, if we order the eigenvectors by their eigenvalues, and assume a connected graph, we have $0=\lambda_1<\lambda_2\leq\cdots\leq\lambda_{n}$ and the first eigenvector is proportional to $\mathbf{\phi}_0=[\sqrt{q_1},\ldots,\sqrt{q_n}]^\top$.  
Hence $\mathbf{\Phi}^\top$ maps the vector $\mathbf{\phi}_0$ to $[\sqrt{q_1+\cdots+q_n},0,\ldots,0]$.  
We define $\mathbf{\Phi}^\top$ to be the elementary block transform of the RA-GFT with inverse $\mathbf{\Phi}$.
\subsection{Relation to other transforms}
\paragraph*{Relation to RAHT.}
As a special case, consider the two-node graph with $\mathcal{V}=\{\mathbf{v}_1,\mathbf{v}_2$, $\mathcal{E}=\{(1,2)\}$, edge weights $w_{12}=w_{21}=1$, and node weights $q_1>0$ and $q_2>0$.  Then  $\mathbf{L}=\left[\begin{smallmatrix}1&-1\\-1&1\end{smallmatrix}\right]$, and $\mathbf{Q} = \diag(q_1, q_2)$, hence the $\mathbf{Q}$-normalized Laplacian is
\begin{eqnarray}
    \mathbf{L}_\mathbf{Q}
    & = & \left[\begin{array}{cc}
        \frac{1}{q_1} & \frac{-1}{\sqrt{q_1q_2}} \\
        \frac{-1}{\sqrt{q_1q_2}} & \frac{1}{q_2}
        \end{array}\right] \\
    & = & \left[\begin{array}{cc}
        a & -b \\
        b & a
        \end{array}\right]
    \left[\begin{array}{cc}
        0 & 0 \\
        0 & \lambda_{\max}
        \end{array}\right]
    \left[\begin{array}{cc}
        a & b \\
        -b & a
        \end{array}\right],
    \label{eqn:lmax}
\end{eqnarray}
where $a={\sqrt{q_1}}/{\sqrt{q_1+q_2}}$,  $b={\sqrt{q_2}}/{\sqrt{q_1+q_2}}$, and $\lambda_{\max}={q^{-1}_1}+{q^{-1}_2}$.  The $2\times 2$ matrix $\mathbf{\Phi}^\top=\left[\begin{smallmatrix}a&b\\-b&a\end{smallmatrix}\right]$ is the RAHT butterfly \cite{queiroz2016compression}.  Hence RAHT is the $2\times 2$ case of the RA-GFT.
\paragraph*{Relation to IAGFT.} 
Consider $\mathbf{Z}=\mathbf{Q}^{-1}\mathbf{L}$, the fundamental matrix of the $(\mathbf{L},\mathbf{Q})$-IAGFT \cite{girault2018irregularity}.  
Clearly $\mathbf{Z}$ is related to $\mathbf{L}_\mathbf{Q}$ by a similarity transform \cite[Remark 1]{girault2018irregularity}: $\mathbf{Z}=\mathbf{Q}^{-1/2}\mathbf{L}_\mathbf{Q}\mathbf{Q}^{1/2}$. 
Thus $\mathbf{Z}$ and $\mathbf{L}_\mathbf{Q}$ have the same  set of eigenvalues, $\mathbf{\Lambda}$,  
and the matrix $\mathbf{U}$ of eigenvectors of $\mathbf{Z}$ is related to $\mathbf{\Phi}$ by $\mathbf{U}=\mathbf{Q}^{-1/2}\mathbf{\Phi}$, that is,
$    \mathbf{Z}
    =\mathbf{U}\mathbf{\Lambda}\mathbf{U}^{-1}
    =\mathbf{Q}^{-1/2}(\mathbf{\Phi}\mathbf{\Lambda}\mathbf{\Phi}^\top)\mathbf{Q}^{1/2}.$
It can be shown \cite[Thm.~1]{girault2018irregularity} that the columns of $\mathbf{U}$ are orthonormal under the $\mathbf{Q}$-norm, i.e., $\mathbf{U}^\top\mathbf{Q}\mathbf{U}=\mathbf{I}$.   The IAGFT is defined as $\mathbf{F}=\mathbf{U}^{-1}$.  So $\mathbf{F}=\mathbf{U}^\top\mathbf{Q}=\mathbf{\Phi}^\top\mathbf{Q}^{1/2}$, or $\mathbf{\Phi}^\top=\mathbf{F}\mathbf{Q}^{-1/2}$.  
Hence the IAGFT is a block transform of the RA-GFT applied to a $\mathbf{Q}^{1/2}$-weighted signal.
%
%
%
%
\subsection{Definition of full RA-GFT}
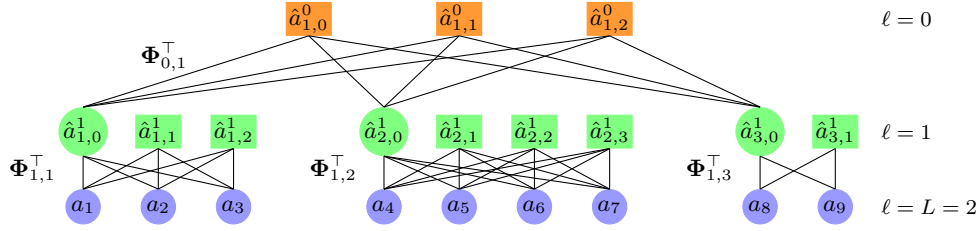
\begin{figure*}[tb]
    \centering
    \begin{tikzpicture}[scale=1]
        \node[blue!40!white,draw,circle,inner sep=1pt,fill,text=black] (a1) at (0, 0) {$a_1$};
        \node[blue!40!white,draw,circle,inner sep=1pt,fill,text=black] (a2) at (1, 0) {$a_2$};
        \node[blue!40!white,draw,circle,inner sep=1pt,fill,text=black] (a3) at (2, 0) {$a_3$};
        
        \node[blue!40!white,draw,circle,inner sep=1pt,fill,text=black] (a4) at (4, 0) {$a_4$};
        \node[blue!40!white,draw,circle,inner sep=1pt,fill,text=black] (a5) at (5, 0) {$a_5$};
        \node[blue!40!white,draw,circle,inner sep=1pt,fill,text=black] (a6) at (6, 0) {$a_6$};
        \node[blue!40!white,draw,circle,inner sep=1pt,fill,text=black] (a7) at (7, 0) {$a_7$};
        
        \node[blue!40!white,draw,circle,inner sep=1pt,fill,text=black] (a8) at (9, 0) {$a_8$};
        \node[blue!40!white,draw,circle,inner sep=1pt,fill,text=black] (a9) at (10, 0) {$a_9$};
                
        \node[anchor=west] (l2) at (10.5,0) {\footnotesize $\ell=L=2$};

        \node[anchor=east] (Q_1_1) at (-0.25,0.5) {$\mathbf{\Phi}_{1,1}^\top$};
        \node[green!50!white,draw,circle,inner sep=0pt,fill,text=black] (a1_1_0) at (0,1) {$\hat{a}^1_{1,0}$};
        \node[green!50!white,draw,rectangle,inner sep=1pt,fill,text=black] (a1_1_1) at (1,1) {$\hat{a}^1_{1,1}$};
        \node[green!50!white,draw,rectangle,inner sep=1pt,fill,text=black] (a1_1_2) at (2,1) {$\hat{a}^1_{1,2}$};
        \draw (a1.north) -- (a1_1_0.south);
        \draw (a1.north) -- (a1_1_1.south);
        \draw (a1.north) -- (a1_1_2.south);
        \draw (a2.north) -- (a1_1_0.south);
        \draw (a2.north) -- (a1_1_1.south);
        \draw (a2.north) -- (a1_1_2.south);
        \draw (a3.north) -- (a1_1_0.south);
        \draw (a3.north) -- (a1_1_1.south);
        \draw (a3.north) -- (a1_1_2.south);
        
        \node[anchor=east] (Q_1_2) at (3.75,0.5) {$\mathbf{\Phi}_{1,2}^\top$};
        \node[green!50!white,draw,circle,inner sep=0pt,fill,text=black] (a1_2_0) at (4,1) {$\hat{a}^1_{2,0}$};
        \node[green!50!white,draw,rectangle,inner sep=1pt,fill,text=black] (a1_2_1) at (5,1) {$\hat{a}^1_{2,1}$};
        \node[green!50!white,draw,rectangle,inner sep=1pt,fill,text=black] (a1_2_2) at (6,1) {$\hat{a}^1_{2,2}$};
        \node[green!50!white,draw,rectangle,inner sep=1pt,fill,text=black] (a1_2_3) at (7,1) {$\hat{a}^1_{2,3}$};
        \draw (a4.north) -- (a1_2_0.south);
        \draw (a4.north) -- (a1_2_1.south);
        \draw (a4.north) -- (a1_2_2.south);
        \draw (a4.north) -- (a1_2_3.south);
        \draw (a5.north) -- (a1_2_0.south);
        \draw (a5.north) -- (a1_2_1.south);
        \draw (a5.north) -- (a1_2_2.south);
        \draw (a5.north) -- (a1_2_3.south);
        \draw (a6.north) -- (a1_2_0.south);
        \draw (a6.north) -- (a1_2_1.south);
        \draw (a6.north) -- (a1_2_2.south);
        \draw (a6.north) -- (a1_2_3.south);
        \draw (a7.north) -- (a1_2_0.south);
        \draw (a7.north) -- (a1_2_1.south);
        \draw (a7.north) -- (a1_2_2.south);
        \draw (a7.north) -- (a1_2_3.south);
        
        \node[anchor=east] (Q_1_3) at (8.75,0.5) {$\mathbf{\Phi}_{1,3}^\top$};
        \node[green!50!white,draw,circle,inner sep=0pt,fill,text=black] (a1_3_0) at (9,1) {$\hat{a}^1_{3,0}$};
        \node[green!50!white,draw,rectangle,inner sep=1pt,fill,text=black] (a1_3_1) at (10,1) {$\hat{a}^1_{3,1}$};
        \draw (a8.north) -- (a1_3_0.south);
        \draw (a8.north) -- (a1_3_1.south);
        \draw (a9.north) -- (a1_3_0.south);
        \draw (a9.north) -- (a1_3_1.south);
                
        \node[anchor=west] (l1) at (10.5,1) {\footnotesize $\ell=1$};
        
        \node[anchor=east] (Q_0_1) at (1.5,2) {$\mathbf{\Phi}_{0,1}^\top$};
        \node[orange!80!white,draw,rectangle,inner sep=1pt,fill,text=black] (a0_1_0) at (3,2.5) {$\hat{a}^0_{1,0}$};
        \node[orange!80!white,draw,rectangle,inner sep=1pt,fill,text=black] (a0_1_1) at (5,2.5) {$\hat{a}^0_{1,1}$};
        \node[orange!80!white,draw,rectangle,inner sep=1pt,fill,text=black] (a0_1_2) at (7,2.5) {$\hat{a}^0_{1,2}$};
        \draw (a1_1_0.north) -- (a0_1_0.south);
        \draw (a1_1_0.north) -- (a0_1_1.south);
        \draw (a1_1_0.north) -- (a0_1_2.south);
        \draw (a1_2_0.north) -- (a0_1_0.south);
        \draw (a1_2_0.north) -- (a0_1_1.south);
        \draw (a1_2_0.north) -- (a0_1_2.south);
        \draw (a1_3_0.north) -- (a0_1_0.south);
        \draw (a1_3_0.north) -- (a0_1_1.south);
        \draw (a1_3_0.north) -- (a0_1_2.south);
        
        \node[anchor=west] (l3) at (10.5,2.5) {\footnotesize $\ell=0$};
    \end{tikzpicture}
    \caption{RA-GFT for point cloud represented by tree structured partition of Figure \ref{fig:tree_cluster}. Circles represent DC coefficients that are further processed by block transforms, while squares contain RA-GFT  coefficients.}
    \label{fig:ra-gft-transform}
\end{figure*}
We now show how the RA-GFT block transforms are composed to form the full RA-GFT.  Consider a list of $N$ points $\mathbf{V}$, a real-valued attribute signal $\mathbf{a}$ and node weights $\mathbf{q}$ on those points:
\begin{equation*}
    \mathbf{V} = 
\begin{bmatrix}
   \mathbf{v}_i
  \end{bmatrix},\quad 
\mathbf{a} = 
\begin{bmatrix}
   {a}_i 
\end{bmatrix},\quad 
\mathbf{q} = 
\begin{bmatrix}
   {q}_i 
\end{bmatrix}.
\end{equation*}
Here, $\mathbf{v}_i$ is an abstract point, and can be considered a node or vertex of a discrete structure.  Now suppose we are given a hierarchical partition or nested refinement of the points, as illustrated in Fig.~\ref{fig:tree_cluster}.  Let $L$ be the depth of the hierarchy, and for each level $\ell$ from the root ($\ell=0$) to the leaves ($\ell=L$), let $\mathbf{v}_i^\ell$ be the $i$th of $M_\ell$ nodes at level $\ell$.  At level $L$, we have $M_L=N$ and $\mathbf{v}_i^L=\mathbf{v}_i$.  For level $\ell<L$, let $\mathcal{C}_i^\ell\subseteq\{1,\ldots,M_{\ell+1}\}$ denote the indices of the children of node $\mathbf{v}_i^\ell$ and let $\mathcal{D}_i^\ell\subseteq\{1,\ldots,N\}$ denote the indices of the descendants of $\mathbf{v}_i^\ell$.  We are also given for all $\ell<L$ and $i\in\{1,\ldots,M_\ell\}$, a graph $\mathcal{G}_i^\ell=(\mathcal{V}_i^\ell,\mathcal{E}_i^\ell,\mathbf{W}_i^\ell,\mathbf{Q}_i^\ell)$, where $\mathcal{V}_i^\ell=\{\mathbf{v}_k^{\ell+1}:k\in\mathcal{C}_i^\ell\}$ is the set of children of node $\mathbf{v}_i^\ell$, $\mathcal{E}_i^\ell$ is a set of edges between the children, $\mathbf{W}_i^\ell$ is a matrix of edge weights, and $\mathbf{Q}_i^\ell=\diag(q_k^{\ell+1}:k\in\mathcal{C}_i^\ell)$ is the diagonal matrix of child node weights, where $q_k^{\ell+1}=\sum_{j\in\mathcal{D}_k^{\ell+1}}q_j$ is the sum of the weights of all $\smash{n_k^{\ell+1}=|\mathcal{D}_k^{\ell+1}|}$ descendants of child $\mathbf{v}_k^{\ell+1}$.  Let $\mathbf{\Phi}_{\ell,i}^\top$ be the RA-GFT block transform of the graph $\mathcal{G}_i^\ell$.

The full $N\times N$ RA-GFT $\mathbf{T}$ is a composition of $N\times N$ orthogonal transforms, $\hat{\mathbf{a}}=\mathbf{T}\mathbf{a}=\mathbf{T}_0\mathbf{T}_1\cdots\mathbf{T}_{L-1}\mathbf{a}$, with inverse  $\mathbf{a}=\mathbf{T}^{-1}\hat{\mathbf{a}}=\mathbf{T}_{L-1}^{-1}\cdots\mathbf{T}_1^{-1}\mathbf{T}_0^{-1}\hat{\mathbf{a}}$, where
\begin{equation}
    \mathbf{T}_\ell=\bdiag(\mathbf{T}_{\ell,1},\ldots,\mathbf{T}_{\ell,M_\ell}),
\end{equation}
is an $N\times N$ orthonormal block diagonal matrix and block $\mathbf{T}_{\ell,i}$ is an $n_i^\ell\times n_i^\ell$ orthonormal matrix for transforming the $n_i^\ell$ descendants of the $i$th node $\mathbf{v}_i^\ell$ at level $\ell$.  Specifically,
\begin{equation}
    \mathbf{T}_{\ell,i}=\mathbf{P}_{\ell,i}^{-1}\bdiag(\mathbf{\Phi}_{\ell,i}^\top,\mathbf{I})\mathbf{P}_{\ell,i}
\end{equation}
where $\mathbf{P}_{\ell,i}$ is an $n_i^\ell\times n_i^\ell$ permutation that collects the lowpass coefficients of the child nodes $\mathbf{v}_k^{\ell+1}$, $k\in\mathcal{C}_i^\ell$ for processing by the RA-GFT block transform $\mathbf{\Phi}_{\ell,i}^\top$ to produce lowpass and highpass coefficients for parent node $\mathbf{v}_i^\ell$.
When the  RA-GFT is implemented using all levels of  the tree, it  produces  a single approximation or low pass (DC) coefficient, and $N-1$ detail or high pass (AC) coefficients. Figure \ref{fig:ra-gft-transform} depicts the application of the RA-GFT for the nested partition depicted in Figure \ref{fig:tree_cluster}. 
It can be seen inductively that $\mathbf{T}$ maps the signal $[\sqrt{q_i}]$ to a single lowpass (DC) coefficient equal to $(\sum_iq_i)^{1/2}$, while all other, highpass (AC), coefficients are equal to 0.  Thus the first, lowpass (DC) basis function of $\mathbf{T}$ is proportional to $[\sqrt{q_i}]$.  In the usual case when $q_i=1$ for all $i$, the first basis function of $\mathbf{T}$ is constant, as desired. This can be verified using Figure \ref{fig:ra-gft-transform}. Assume the attributes $a_i$ and  weights $q_i$ are all equal to $1$. Since the block transforms at level $\ell=1$ have sizes $3$, $4$ and $2$, the only non zero coefficients at level $\ell=1$ are $\hat{a}^1_{1,0}= \sqrt{3}$,  $\hat{a}^1_{2,0}= \sqrt{4}$ and  $\hat{a}^1_{3,0}= \sqrt{2}$. Then at level $\ell=0$, the weight matrix is $\mathbf{Q}_0^0 = \diag(3, 4, 2)$, hence the first column of $\mathbf{\Phi}_{0,1}$ is  $(1\sqrt{9})[\sqrt{3},\sqrt{4}, \sqrt{2}]^{\top}$. Then the only nonzero transform coefficient produced at level $\ell =0$ is $\hat{a}_{1,0}^0 = \sqrt{9}$. Therefore the RA-GFT of the $\mathbf{a=1}$  vector is a $1$-sparse vector.
\begin{table*}[ht]
\centering
\scalebox{1}
{
\begin{tabular}{|c|c|c|c|c |c |}
\hline 
     RA-GFT $b_L=2$    &   RA-GFT $b_L=4$   &   RA-GFT $b_L=8$  &   RA-GFT $b_L=16$ &   block-GFT $b=8$    &    block-GFT $b=16$ \\ \hline
           $0.2098 \times 10^3$    &  $0.5243 \times 10^3$          &  $6.12 \times 10^3$           & $104.25 \times 10^3$ 
           & $5.97 \times 10^3$  & $104.13 \times 10^3$ \\
     \hline 
\end{tabular}
}
\caption{ Complexity estimate of the RA-GFT and block-GFT as a function of the block size.    }
\label{tab_complexity}
\end{table*}
\section{Application to Point Clouds}
\label{sec:pointclouds}
In a point cloud, the vertices $\mathbf{V}=[\mathbf{v}_i]$ represent the coordinates $(x,y,z)$ of real points in space;  the attributes $\mathbf{a}=[a_i]$ represent colors or other attributes of the points; and the weights $\mathbf{q}=[q_i]$ represent the relative importance of the points.  The weights are usually set to be constant ($q_i=1$), but may be adjusted to reflect different regions of interest \cite{sandri2019roi}.
%
 %
We assume points are voxelized. A voxel is a volumetric unit of the domain of a $3$-dimensional signal, analogous to a pixel in the $2$-dimensional case. Let $J$ be a positive integer, and partition the space into $2^J\times2^J\times2^J$ voxels.  We say a point cloud is voxelized with depth $J$ if all the point coordinates take values in the integer grid $\{ 0, 1, \ldots, 2^J-1 \}$.
A voxelized point cloud can be organized into a hierarchical structure. The process is described in \cite{queiroz2016compression,pavez2018dynamic}.
The voxel space $\{0,1,\ldots,2^J-1\}^3$ is hierarchically partitioned into sub-blocks of size $b_{\ell} \times b_{\ell} \times b_{\ell}$, where $b_{\ell}$ is a power of $2$. These block sizes allow for a hierarchy with $L$ levels, where $\prod_{\ell =1}^L b_{\ell} = 2^J$. Levels are ordered according to resolution.
The partition is constructed by generating a point cloud for each resolution level.   That is, beginning with $\mathbf{V}^L=\mathbf{V}$, we have for $\ell<L$:
\begin{equation}\label{eq_pointcloud_point_hierarchy}
     \mathbf{V}^{\ell} = \unique\left(\floor\left(\frac{ \mathbf{V}^{\ell+1}}{b_{\ell+1}} \right) \right),
\end{equation}
where the $\unique$ function removes points with equal coordinates.
%
Each point $\mathbf{v}_i^\ell$ in $\mathbf{V}^\ell$ has children
\begin{equation}\label{eq_pointcloud_parent_children}
     \mathcal{V}_i^{\ell} = \left\lbrace \mathbf{v}_k^{\ell+1} : \mathbf{v}_i^{\ell} =  \floor\left(\frac{ \mathbf{v}_k^{\ell+1}}{b_{\ell+1}} \right) \right\rbrace.
\end{equation}
With the children we form a graph $\mathcal{G}_i^\ell=(\mathcal{V}_i^\ell,\mathcal{E}_i^\ell,\mathbf{W}_i^\ell,\mathbf{Q}_i^\ell)$, where there is an edge between nodes $\mathbf{v}_j^{\ell+1},\mathbf{v}_k^{\ell+1}\in\mathcal{V}_i^\ell$ if the distance between $\smash{\mathbf{v}_j^{\ell+1}}$ and $\smash{\mathbf{v}_k^{\ell+1}}$ is less than a threshold, in which case the weight $w_{jk}$ is set to a decreasing function of that distance.  Using this hierarchy and set of graphs, the RA-GFT is constructed and applied to the attributes.
The point hierarchies described in (\ref{eq_pointcloud_point_hierarchy}) and (\ref{eq_pointcloud_parent_children}) can be obtained in $\mathcal{O}(N\log N)$ time  \cite{queiroz2016compression,pavez2018dynamic}.  At resolution $\ell +1$ we need to construct $M_{\ell} = \mathcal{O}(N)$ block transforms. For constant block sizes    ($b_{\ell} = \mathcal{O}(1)$), each  block transform can be obtained and applied  in $\mathcal{O}(1)$ time.  Since there are $L=\mathcal{O}(\log N)$ resolution levels, and  each level is $\mathcal{O}(N)$,    the  RA-GFT has complexity $\mathcal{O}(N\log N)$.
\begin{figure}[ht]
    \centering 
    \begin{subfigure}[b]{0.95\linewidth}
        \centering
        \includegraphics[width=0.9\textwidth]{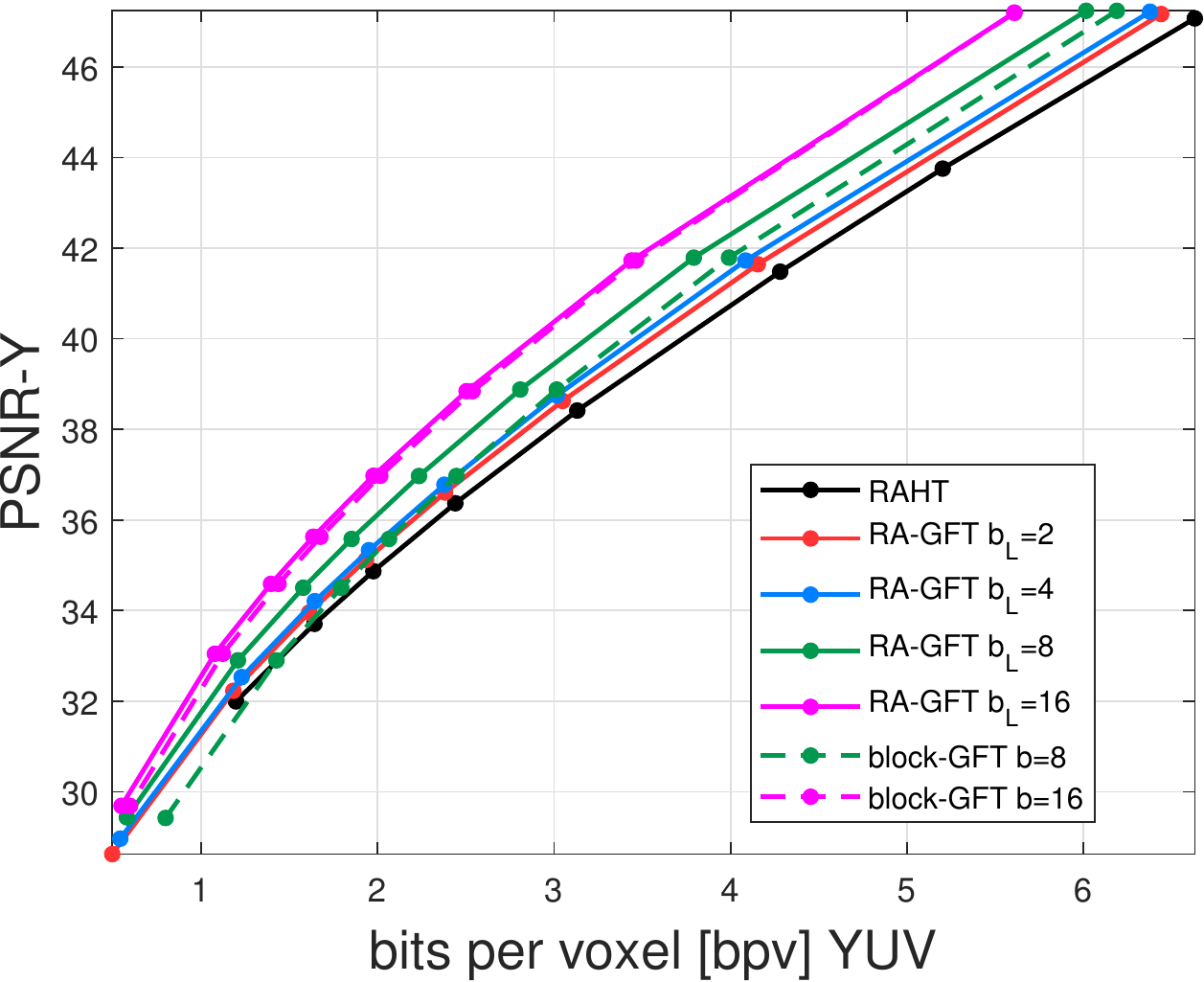}
        \caption{``longdress" sequence}
    \end{subfigure}
    \begin{subfigure}[b]{0.95\linewidth}
        \centering
        \includegraphics[width=0.9\textwidth]{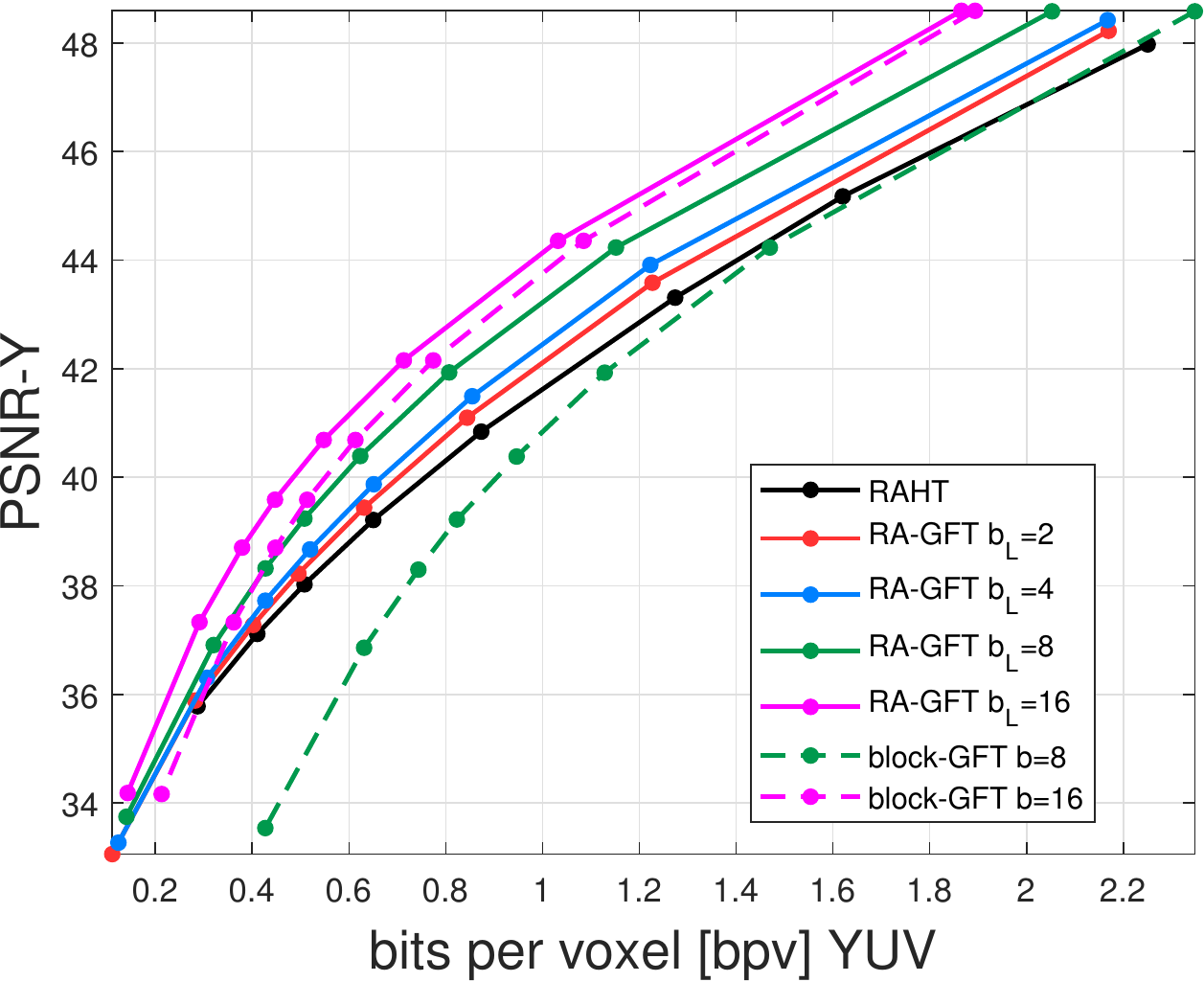}
        \caption{``loot" sequence}
    \end{subfigure}
    \caption{Distortion rate curves for color compression.}
    \label{fig:vs_RAHT_vs_blockGFT}
\end{figure}
\section{Experiments}
\label{sec_exp}
In this section, we evaluate the RA-GFT in compression of color attributes of the ``8iVFBv2'' point cloud dataset\footnote{\url{https://jpeg.org/plenodb/pc/8ilabs/}} \cite{d20178i}, which consists of four sequences: ``longdress'', ``redandblack'', ``loot'' and ``soldier'',  
and compare its performance to  block-GFT \cite{zhang2014point} and RAHT \cite{queiroz2016compression}. 
Colors are transformed from RGB to YUV spaces, and each of the Y, U, and V components are processed independently  by the transform. 
For all transforms, we perform uniform quantization and entropy code the coefficients using the adaptive run-length Golomb-Rice algorithm \cite{malvar2006adaptive}. Distortion   for the Y component is given by
\begin{equation*}
            PSNR_Y = -10\log_{10} \left(\frac{1}{T}\sum_{t=1}^T \frac{\Vert \mathbf{Y}_t - \hat{\mathbf{Y}}_t  \Vert_2^2}{255^2 N_t}  \right),
\end{equation*}
where $T=300$ is the number of frames in the sequence,  $N_t$ is the number of points in the $t$-th frame, and $\mathbf{Y}_t$ and $\hat{\mathbf{Y}}_t$ represent the original and decoded signals of frame $t$.  Rate is reported in bits per voxel [bpv]  $R = \left({\sum_{t=1}^T B_t}\right)/\left({\sum_{t=1}^T N_t}\right),$
where $B_t$ is the number of bits used to encode the YUV components of the $t$-th frame.
\subsection{Compression of color attributes}
Each point cloud in the 8iVFB dataset is voxelized with depth $J=10$. Therefore  $L \leq 10$ and its value will  depend on the  block sizes. We implement several RA-GFTs, each with a different block size at the highest resolution (level $L$),  but with the same block sizes $b_{\ell} =2$ for  all other resolutions ($\ell <L$). When $b_L$ is equal to $2$, $4$, $8$, and $16$,   $L$ is equal to $10$, $9$, $8$ and $7$ respectively. For the block-GFT we choose block sizes $b=8$ and $b=16$. Graphs are constructed by adding edges if the distance between a pair of point coordinates is below a fixed  threshold, while edge weights are  the reciprocal of the distance,  as in \cite{zhang2014point}.   Distortion rate curves  are shown in Figure~\ref{fig:vs_RAHT_vs_blockGFT}.

The RA-GFT provides substantial gains over RAHT. When the block size is smallest $b_L=2$, the corresponding RA-GFT outperforms RAHT up to $0.5$db for the ``longdress'' sequence, and up to $1$db for the ``loot'' sequence. Similar results were obtained for other sequences, not shown due to lack of space. Coding performance improves as the block size $b_L$ increases, up to $2.5$ dB over RAHT on both sequences. The block-GFT also has this property, however, it requries a large block size ($b=16$) to consistently outperform the RAHT for all sequences. This could occur because for smaller blocks, the block-GFT DC coefficients may still be highly correlated. Since the RA-GFT with small blocks can be viewed as an extension of  the block-GFT to multiple levels, the transform coefficients of the proposed transform are less correlated.  
%
%
\subsection{Complexity analysis}
 At  each  level of RA-GFT, multiple GFTs of different sizes are constructed, so that the overall computational complexity is dominated by the number and size of those transforms. 
At resolution level $\ell$, the $i$th transform is a  $\vert \mathcal{V}_i^{\ell} \vert \times \vert \mathcal{V}_i^{\ell} \vert$ matrix, which  is obtained by eigendecomposition in  roughly $\vert \mathcal{V}_i^{\ell} \vert^3$ operations. 
As a proxy for the total number of operations required by the  RA-GFT we use
\begin{equation}\label{eq_proxy_complexity}
    K = \sum_{\ell =0}^{L-1} \sum_{i =1}^{M_{\ell}} \vert \mathcal{V}_i^{\ell} \vert^3.
\end{equation}
 Eq. (\ref{eq_proxy_complexity}) is consistent with Section \ref{sec:pointclouds}, since 
 $\vert \mathcal{V}_i^{\ell} \vert^3 = \mathcal{O}(1)$, $M_{\ell} = \mathcal{O}(N)$ and $L \leq \log(N)$.
We consider a collection of $T$ point clouds. The $t$-th point cloud has $N_t$ points,  and  the quantity (\ref{eq_proxy_complexity}) for the RA-GFT on that point cloud   is denoted by $K_t$. The complexity proxy for the RA-GFT with a given tree structured nested partition is defined as $C = (\sum_{t=1}^T K_t)/(\sum_{t=1}^T N_t)$. We compute this quantity for the RA-GFT and block-GFT depicted in Figure~\ref{fig:vs_RAHT_vs_blockGFT}. Our results are shown in Table~\ref{tab_complexity}, and  are computed from  the first $10$ point clouds of each sequence of the ``8iVFB'' dataset, for a total of $T=40$ point clouds.  For $b_L=8$, the increase in complexity from the block-GFT to the RA-GFT is only 2.5\%; for $b_l=16$ the increase is negligible. More importantly, for smaller blocks, the complexity of the RA-GFT is orders of magnitude lower than of block-GFT. 
\section{Conclusion}
\label{sec_conclusion}
By allowing multiple block sizes, and multiple levels of resolution, the  proposed RA-GFT  can be viewed as a generalization of  the block-GFT and the RAHT, reaching coding efficiency  comparable to block-GFT, with  computational complexity slightly higher than RAHT. Thefore, the RA-GFT achieves better   performance-complexity trade-offs than these transforms.
 By using a non-separable transform on  larger blocks  the RA-GFT can exploit local geometry more efficiently than the RAHT. By applying transforms with small blocks at  multiple resolutions, the RA-GFT can  approach the performance of the block-GFT with a reduced complexity.  In addition, for large transform blocks at resolution $L$, the RA-GFT outperforms the block-GFT, with a negligible complexity increase. 
%
%
\vfill
\pagebreak
\bibliographystyle{IEEEbib}
\bibliography{refs}

\end{document}